\pdfoutput=1

\documentclass[11pt]{article}

\usepackage[final]{acl}

\usepackage{times}
\usepackage{latexsym}
\usepackage{pdfpages}
\usepackage{wrapfig}
\usepackage{caption}
\usepackage{adjustbox}

\usepackage[utf8]{inputenc} 
\usepackage[T1]{fontenc}    
\usepackage{hyperref}       
\usepackage{url}            
\usepackage{booktabs}       
\usepackage{amsfonts}       
\usepackage{nicefrac}       
\usepackage{microtype}      
\usepackage{xcolor}         

\usepackage{subcaption}
\usepackage{threeparttable}
\usepackage{commath}
\usepackage{amsthm}
\usepackage{amsfonts}
\usepackage{multirow}
\usepackage{booktabs}
\usepackage{xcolor}
\usepackage{float}
\usepackage{amsmath}
\usepackage{amssymb}
\usepackage{bbm}
\definecolor{darkgreen}{RGB}{0, 100, 0}

\theoremstyle{definition}
\newtheorem{definition}{Definition}[section]
\usepackage[linesnumbered,ruled]{algorithm2e}

\newcommand{\ModalSCM}{CodeSCM}
\newcommand{\MultiModal}{multi-modal}

\definecolor{suman}{HTML}{FF0000}

\usepackage[T1]{fontenc}

\usepackage[utf8]{inputenc}

\usepackage{microtype}

\usepackage{inconsolata}

\usepackage{graphicx}

\title{\ModalSCM{}: Causal Analysis for Multi-Modal Code Generation}

\author{%
  Mukur Gupta$^*$
  \hspace{0.25em}
  \hspace{0.25em}
  Noopur Bhatt$^*$
  \hspace{0.25em}
  \hspace{0.25em}
  Suman Jana \\
  Columbia University \\
  \texttt{\{mukur.gupta, noopur.bhatt\}@columbia.edu} \\
  \texttt{suman@cs.columbia.edu} \\
}

\begin{document}
\maketitle
\def\thefootnote{*}\footnotetext{These authors contributed equally to this work.}\def\thefootnote{\arabic{footnote}}

\begin{abstract}
In this paper, we propose \ModalSCM{}, a Structural Causal Model (SCM) for analyzing \MultiModal{} code generation using large language models (LLMs). By applying interventions to \ModalSCM{}, we measure the causal effects of different prompt modalities, such as natural language, code, and input-output examples, on the model. \ModalSCM{} introduces latent mediator variables to separate the code and natural language semantics of a \MultiModal{} code generation prompt. Using the principles of Causal Mediation Analysis on these mediators we quantify direct effects representing the model's spurious leanings. We find that, in addition to natural language instructions, input-output examples significantly influence code generation.
\end{abstract}

\section{Introduction}
\label{sec:introduction}
Modern Large Language Models (LLMs) have shown remarkable effectiveness in code reasoning tasks, particularly code generation \cite{nijkamp2023codegen, rozière2023code, qwen}. This task involves generating code that meets specific \MultiModal{} requirements, constrained by natural language instructions, code snippets, and input-output (I/O) example pairs \cite{humaneval, austin2021program}. Additionally, some \MultiModal{} prompt components contain information from both code and natural language modalities~\cite{dual_channels}, such as function signatures and variable names, where code structure and natural language coexist. This enriched coding context, combining programming and natural language semantics, helps LLMs better understand both the semantics and syntactic requirements of the desired code.

Prior research has shown the effectiveness of prompt tuning in improving generation performance~\cite{brown2020language, liu2021pretrain, wei2023chainofthought}. 
These works have shown that \MultiModal{} prompts can be highly sensitive, where small adjustments might result in drastically different responses from the model~\cite{chao2023jailbreaking, zhu2023promptbench, sclar2023quantifying}. However, the interactions between the \MultiModal{} components of code generation prompts and their direct or indirect effects on the generated code are still not well understood.

\begin{figure*}[t]
    \centering
    \includegraphics[width=0.93\textwidth]
    {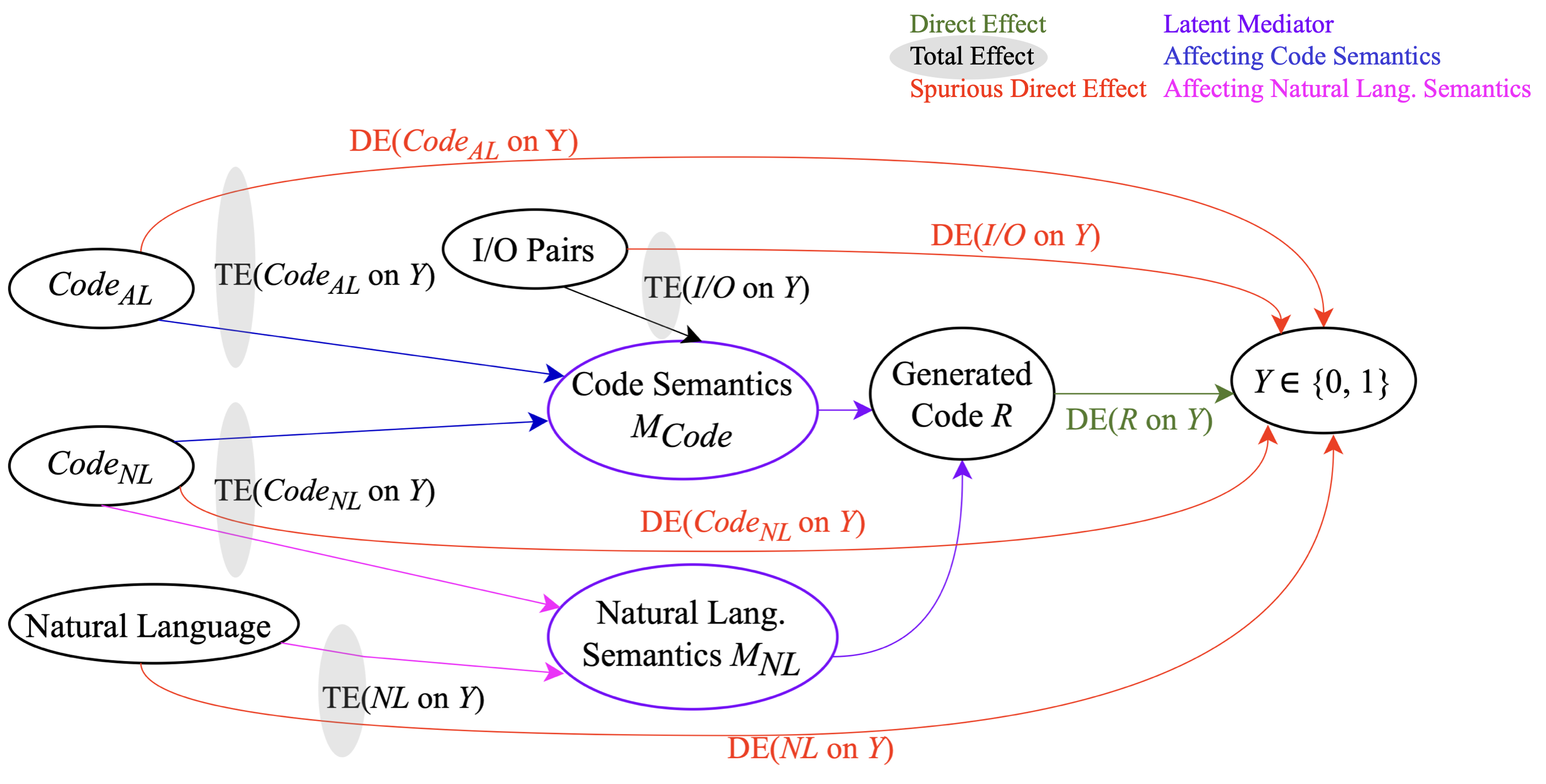}
    \caption{\ModalSCM{} causal graph representing the total and direct effects of the modal variable nodes on the response variable $Y$  representing the correctness of the generated code. $Code_{AL}$ represents algorithmic channel of code and $Code_{NL}$ is natural language channel of code.}
    \label{fig:causal_graph}
    \vspace{-15pt} 
\end{figure*}

In this paper, we systematically explore these complex \MultiModal{} effects using a causal approach. We propose a novel causal framework, \ModalSCM{}, to measure the causal effects of different modalities in the prompt on the performance of code generation LLMs. \ModalSCM{} defines a Structural Causal Model\cite{pearl2000models}, shown in Figure~\ref{fig:causal_graph},  where each modal component of the prompt is treated as an independent variable that causally affects the code generated by the model. To account for similar natural language and code semantics of different surface forms, we introduce two latent mediator variables to capture code semantics and natural language semantics of the input prompt, mimicking a human mental model for generating correct code snippets from a multi-modal input problem.\\
\indent Specifically, we make four key contributions in this paper: (i) we introduce \ModalSCM{}, a novel framework for causal inference in \MultiModal{} code generation tasks, enhancing interpretability and causal understanding of codeLLMs. While \ModalSCM{} is designed for the code generation task in this paper, it can be extended to other modalities, tasks, and transformations. (ii) using \ModalSCM{}, we define the Total Effects of the modalities on code generation, highlighting that input-output example pairs and natural language code components, like function headers, are significant modal components alongside natural language instructions. We also observe benchmark memorization in LLMs like GPT-4T with our Total Effect analysis; (iii) through targeted interventions on \ModalSCM{}, we measure the Direct Effects of each modality, representing spurious model correlations, and show that simple semantics-preserving transformations to input-output example pairs lead to a significant drop in accuracy; and (iv) following the asymmetric causal effects of modalities, we examine the effect of \MultiModal{} code-specific pretraining on the embedding space, which shows that codeLLM CodeLLaMa can align different prompt modalities better than LLaMa-2 in the embedding space. Our code is available on GitHub\footnote{\href{https://github.com/nb15/codeSCM-naacl25}{https://github.com/nb15/codeSCM-naacl25}}.

\section{Background}
\label{background}
\subsection{Structural Causal Model}
\label{background:scm}
A Structural Causal Model (SCM) $\mathcal{M}$ \cite{pearl2000models} is defined by a 4-tuple 
$\mathcal{M} = \langle \textbf{U}, \textbf{V}, \textbf{F}, \textit{P}(\textbf{U}) \rangle$, where \textbf{U} represents a set of exogenous variables that are affected by factors external to the model, $P(\textbf{U})$ is a joint probability distribution over the set \textbf{U}, \textbf{V} is a set of endogenous variables determined by variables in \textbf{U}$\cup$\textbf{V},  and \textbf{F} is a set of functions from \textbf{U}$\cup$\textbf{V} to \textbf{V}. 
The functions in \textbf{F} can range from simple indicator functions for binary variables to language models for complex NLP tasks. The SCM $\mathcal{M}$ can be represented by a causal graph $\mathcal{G}$, which employs nodes to represent both exogenous and endogenous variables.

Causal effects on any response variable $Y\in\textbf{V}$ are quantitatively measured using interventions. An intervention on $X\in \textbf{V}$, represented by $do(x)$, creates a sub-SCM $\mathcal{M}^x = \langle \textbf{U}, \textbf{V}, \textbf{F}_x, \textit{P}(\textbf{U}) \rangle$ where $\textbf{F}_x$ represents a subset of function mappings in \textbf{F} which do not have $X$ in their co-domain, and $X$ is replaced by a constant $x$. Hence, forcing the variable $X$ to take a constant value and removing all mechanisms that may affect it. The response variable $Y$ post-intervention ($do(x)$) is represented as $Y_x$, i.e., $P(Y_x) = P(Y|do(x))$. Following the above notation, we can formally define causal effects:
\begin{definition}{\label{def:TE}\textbf{(Total Effect)}~\cite{pearl2000causality}:} 
The causal effect of two distinct realizations of variable $X$ with $do(X = x^{\prime})$ and $do(X = x^{\prime\prime})$. Total Effect ($TE (x^{\prime}, x^{\prime\prime})$) can be written as:
\begin{equation}
\mathbb{E}[Y | do(X=x^{\prime})] - \mathbb{E}[Y | do(X=x^{\prime\prime})]
\end{equation}
\end{definition}

Causal Mediation Analysis \cite{robins1992identifiability, pearl2022direct, robins2003semantics} involves understanding the effects of a mediator $M\in \textbf{V}$ in explaining changes in $Y$. 
All the effects from $X$ to $Y$ where all $Z\in\textbf{V}$, representing the parents of $Y$ excluding $X$, remain fixed are called Direct Effects. 
We measure the direct effect of modalities to define spurious learnings that are not mediated by the latent mediators. We use the definition of Path Effect for systematic measurement of direct effects of a modality.
\begin{definition}{\label{def:PE}\textbf{(Path Effect)}~\cite{avin2005identifiability, wu2019pcfairness}:} 
The causal effect of variable $X$ along a path $\alpha$ can be represented in an edge subgraph $\mathcal{G}_\alpha$. Path Effect ($PE_\alpha (x^{\prime}, x^{\prime\prime})$) can be written as:
\begin{equation}
\mathbb{E}\left[Y | Z_{\alpha}(do(x^{\prime\prime})), Z_{\overline{\alpha}}(do(x^{\prime}))\right]
- \mathbb{E}\left[Y(x^{\prime})\right]
\end{equation}

where $Z_{\alpha}$ is the set of all mediators $\in \mathcal{G}_\alpha$, and $Z_{\overline{\alpha}}$ is the complementary mediator set. Hence, all the variables on the path $\alpha$ take values with $do(x^{\prime\prime})$, and other mediators that do not lie on $\alpha$ take values with $do(x^{\prime})$. Note that the Direct Effect is a special case of the Path Effect.
\end{definition}

\begin{figure*}[!t]
    \centering
    \begin{subfigure}{0.435\linewidth}
        \centering
        \includegraphics[width=\linewidth]{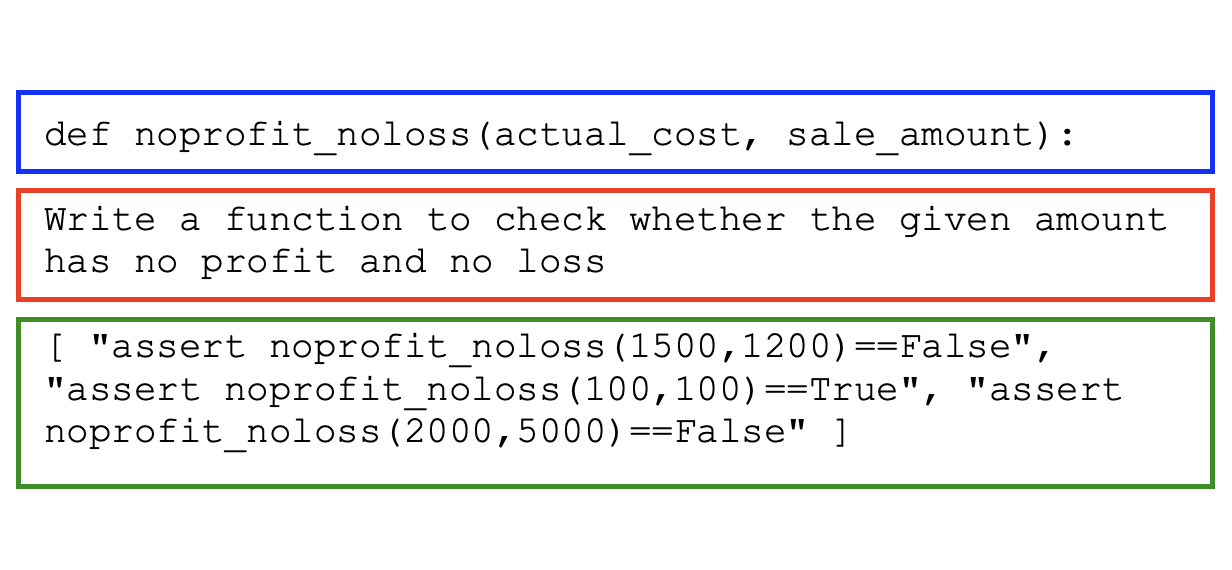}
        \caption{\label{fig:prompt_example}Multi-Modal Prompt}
    \end{subfigure}
\hfill
    \begin{subfigure}{0.54\linewidth}
        \centering
        \includegraphics[width=\linewidth]{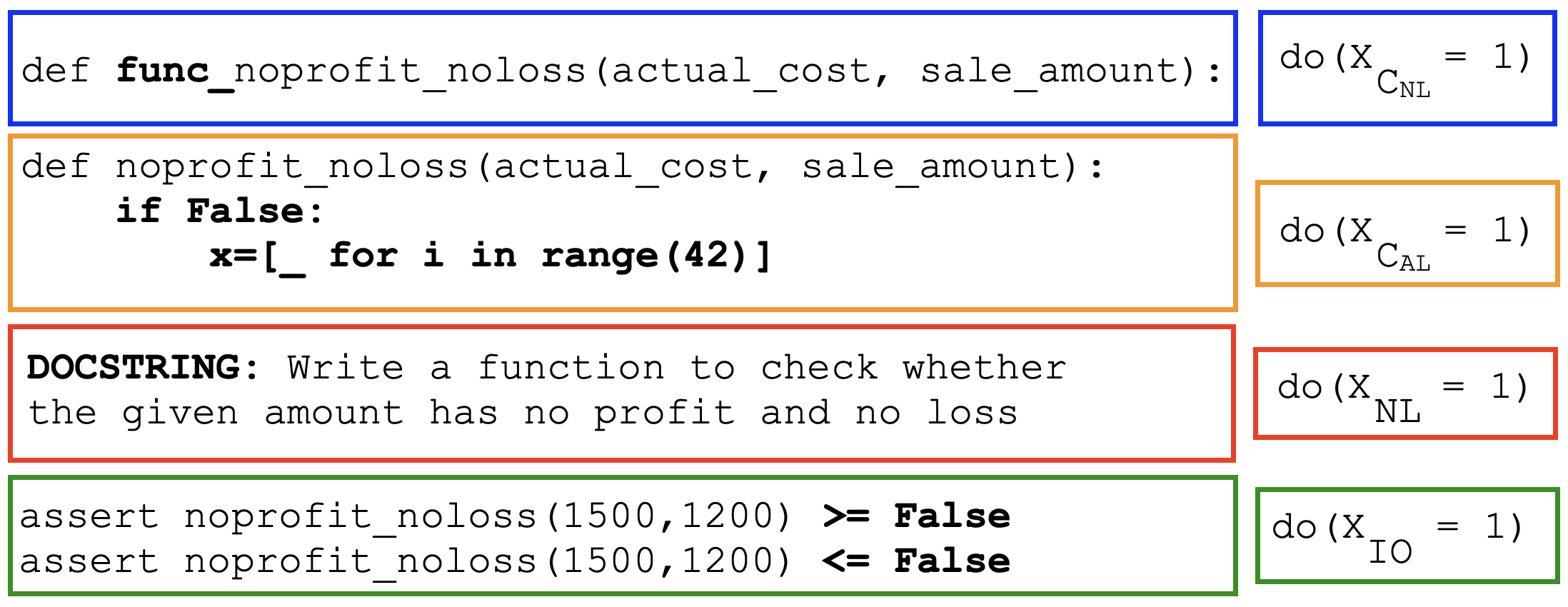}
        \caption{\label{fig:modal_transformations}Semantics preserving transformations}
    \end{subfigure}

\caption{\label{fig:prompt_mbpp_637}(a) Modalities in an example from mMBPP+ dataset, \textcolor{red}{red} for NL, \textcolor{blue}{blue} for $Code_{AL}$ and $Code_{NL}$, and \textcolor{darkgreen}{green} for Input/Output examples. (b) semantics preserving transformations: \textcolor{red}{red} for natural language, \textcolor{blue}{blue} for $Code_{NL}$, \textcolor{orange}{orange} for $Code_{AL}$, and \textcolor{darkgreen}{green} for I/O examples.}
\vskip -0.2in
\end{figure*}

\begin{figure}[h]
    \centering
    \includegraphics[width=0.48\textwidth]{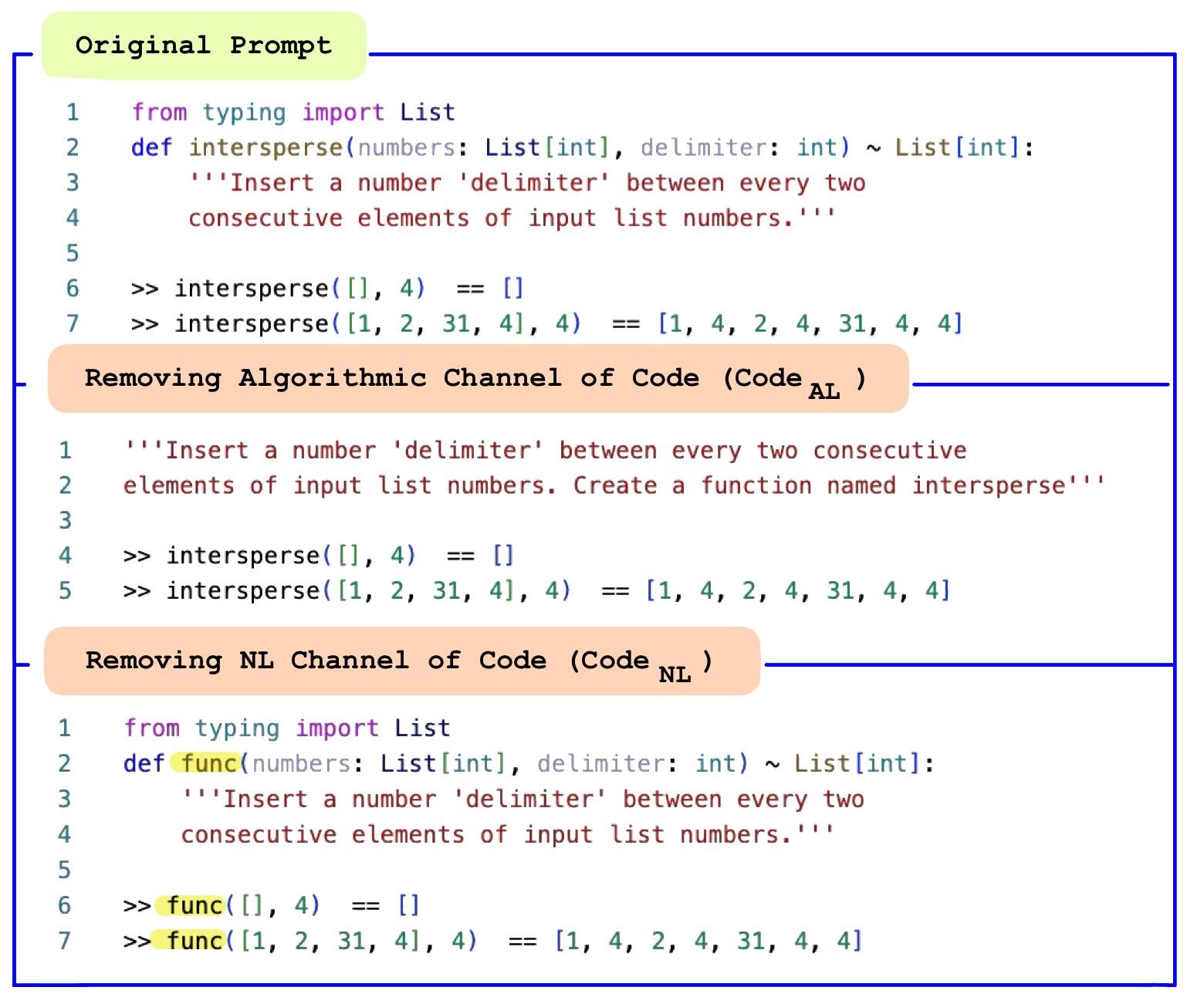}
    \caption{\label{fig:appx_img}The original HumanEval+ prompt (top) includes the function header \texttt{intersperse}, followed by natural language instructions and input-output pairs for code generation. The first modification (middle) removes the algorithmic code channel by eliminating all code components while retaining a natural language description of the function header in the docstring. The second modification (bottom) removes the natural language channel by standardizing the function name.}
\vspace{-15pt} 
\end{figure}

\subsection{Multi-Modal Code Generation}
\label{background:code modality}
For code generation tasks, natural language instructions alone are often insufficient to meet strict context-based syntax requirements, such as variable or function names that are dependent on surrounding code. Thus, natural language prompts are augmented with code modality, guiding the generation into appropriate syntactical space \cite{humaneval, austin2021program}. Additionally, some prompt components, such as function header name, appear as a single entity but contribute to multiple modalities in terms of model understanding, as observed by \citet{dual_channels}. 

For instance, the function header name in Figure~\ref{fig:prompt_example} is primarily a code component, but its natural language name also conveys information about the desired output. 
As highlighted later in Section~\ref{setup:modal scm}, we call this a natural language channel of Code. Figure~\ref{fig:appx_img} illustrates an example of a prompt with and without the natural language channel of Code. Similarly, input-output (I/O) example pairs also carry information about code correctness and logic beyond the syntactical structure. We believe that future codeLLMs might rely heavily on these components to ensure the correctness of intermediary variables, akin to code debugging process of longer code fragments. Therefore, we consider I/O pairs and natural language channels of code as separate modalities, in addition to natural language instructions and code.

\section{Problem Setup}
\subsection{CodeSCM}
\label{setup:modal scm}
Each prompt $\mathcal{P}$ in dataset $\mathcal{D}$ is decomposed into its \MultiModal{} components, which are represented as variables in the structural causal model, as shown in Figure~\ref{fig:causal_graph}. We use the extended Backus–Naur form (Equation~\ref{eq:mm_prompt}) to represent the \MultiModal{} prompt. In \ModalSCM{}, as shown in Equation~\ref{eq:mm_prompt}, we consider four modalities: Natural Language ($NL$), algorithmic channel of Code ($Code_{AL}$), natural language channel of Code ($Code_{NL}$), and input-output example pairs ($I/O$). 

We define the \MultiModal{} structural causal model (\ModalSCM{}) to model the causal relationship between prompt modalities and the model-generated code. Since different code snippets and similar natural language texts can convey the same semantics for a human mental model, we introduce two latent mediators: $M_{Code}$ for code semantics and $M_{NL}$ for natural language semantics. Following the Causal Mediation Analysis, we assume each modality’s effect on the output is mediated through these variables. As shown in Figure~\ref{fig:causal_graph}, $Code_{AL}$ and $NL$ directly affect $M_{Code}$ and $M_{NL}$ respectively; $I/O$ affects $M_{Code}$, and, $Code_{NL}$ directly affects both mediators. The output generated code, $R$, is tested for correctness against the test cases, where code correctness is the response variable $Y \in {0,1}$, with $\mathbb{E}(Y)$ representing accuracy over dataset $\mathcal{D}$. We do not account for confounding variables in this analysis, leaving this investigation for future work.

\subsection{Modal Causal Effects}
\label{sec:causal effects defn}
Using \ModalSCM{}, we define the causal effects of each modal variable in $\mathcal{P}$ on the generated code. We measure the Total Effect (Definition~\ref{def:TE}) of each modality on the response variable $Y$, reflecting the model's sensitivity. Additionally, we examine the Direct Effect of modalities in form of Path Effect (Definition~\ref{def:PE}) on $Y$, along a path that bypass $M_{Code}$ and $M_{NL}$, capturing spurious correlations learned during training. We also define additional variables and interventions to quantify these effects. Direct Effect (DE) and Total Effect (TE) for $Code_{AL}$ are presented here and the detailed derivations for other modalities are in Appendix~\ref{app:causal_effects_extended}.

\paragraph{\textbf{Causal effects of \( \boldsymbol{Code_{AL}}. \)}}
We consider the $Code_{AL}$ variable as an output of a structural equation $F_C \in \textbf{F}$ on three additional variables, $C_{AL}$, $C_{DC}$ and $X_{AL}$, shown in Equation~\ref{eq:code intervention}. $C_{AL}$ is the actual prompt component $\mathcal{P}_{Code_{AL}}\in \mathcal{D}$. To measure $DE(Code_{AL} \ on \ Y)$, quantifying the spurious correlations, we vary $Code_{AL}$ variable while keeping mediator $M_C$ constant i.e $M_C(Code_{AL}) = M_C(Code^{\prime}_{AL})$. As shown in Figure~\ref{fig:modal_transformations}, we do this by inserting Dead Code (DC) into the original code, a code semantics-preserving transformation. The dead code is represented by variable $C_{DC}$. We use the categorical variable $X_{AL}$ to represent the interaction between the actual code and the dead code, which also allows us to perform an intervention and calculate causal effects over all prompts in $\mathcal{D}$. We drop the $AL$ subscript in the following derivation for brevity. It can take one of three values as shown in equation~\ref{eq:code intervention}.
\begin{multline}
\label{eq:code intervention}
    Code_{AL} \gets \mathbbm{1}_{\{ X = 1 \}}(C_{AL} + C_{DC})\\ + \mathbbm{1}_{\{ X = 0 \}}(C_{AL})
    + \mathbbm{1}_{\{ X = -1 \}}({NULL})
\end{multline}
where $\mathbbm{1}(.)$ is an indicator function; $(C_{AL} + C_{DC})$ represents the concatenation of a snippet of dead code with the actual code.

We measure the TE of $Code_{AL}$, $TE(Code_{AL}\ on\ Y)$, by computing the expected change in the $Y$ by setting the $Code_{AL}$ component as NULL in the prompt using the variable $X_{AL}$. 
\begin{align*}
    TE &= TE(do(X_{AL}=0), do(X_{AL}=-1))\\
    &\stackrel{(i)}{=} \mathbb{E}\left[ Y_{X=0} \right] -\mathbb{E}\left[ Y_{X=-1} \right]\\
    &\stackrel{(ii)}{=} \mathbb{P}\left[ Y_{X=0}=1 \right] - \mathbb{P}\left[ Y_{X=-1}=1 \right]\\
    &\stackrel{(iii)}{=}A(\mathcal{D}) - A(\mathcal{D}; \mathcal{P}_{Code_{AL}} =\ NULL)
\end{align*}
where, equality $(i)$ follows from the Definition~\ref{def:TE}, equality $(ii)$ follows because $Y$ follows Bernoulli distribution; $A(\mathcal{D})$ is the accuracy of the model over the dataset $\mathcal{D}$. Figure~\ref{fig:appx_img} shows an example of how $Code_{AL}$ is removed from the prompt for computing the total effect.

The DE of $Code_{AL}$ on $Y$, $DE(Code_{AL} \ on \ Y)$ is measured by the expected change in $Y$ with varying $Code_{AL}$ while keeping $M_{Code}$ unchanged with dead code insertion. We calculate DE using the Path Effect of $X_{AL}$ on $Y$, along a path from $X_{AL}$ to $Y$ which goes through $Code_{AL}$ but skips $M_{Code}$. We note that the quantification of direct effect is a special case path effect definition. Using Definition~\ref{def:PE}:
\begin{align*}
    DE &= \mathbb{E}\left[Y_{Code_{AL}(X=1), M_{C}(X=0)}\right] - 
    \mathbb{E}\left[Y_{X=0}\right]\\
    &\stackrel{(i)}{=}\mathbb{E}\left[Y_{X=1}\right] - \mathbb{E}\left[Y_{X=0}\right]\\
    &\stackrel{(ii)}{=}A(\mathcal{D}) - A(\mathcal{D}; \mathcal{P}_{Code_{AL}} = C_{AL} + C_{DC})
\end{align*}

where equality $(i)$ follows from the fact that $M_{C}(Code_{AL}(X = 0))$ is equal to $M_{C}(Code_{AL}(X = 1))$, because the dead code insertion in Equation~\ref{eq:code intervention} keeps the code semantics $M_{Code}$ unchanged. Equality $(ii)$ is similar to equalities $(ii)$ and $(iii)$ used in TE.

\paragraph{Causal Effects of Other Modalities.}
Similarly, the $NL$, $I/O$, and $Code_{NL}$ modal variables are considered as outputs of structural Equations~\ref{eq:nl intervention}, \ref{eq:io intervention} and \ref{eq:dual intervention} respectively. For $Code_{NL}$, the direct effect requires bypassing two mediators, $M_{NL}$ and $M_{Code}$. Therefore, we define a transformation that preserves semantics for both. As seen in Figure~\ref{fig:modal_transformations}, our transformation adds a prefix DN (Dead Name) to the function header, preserving semantics in both the natural language and code domains. For $I/O$ transformations, each assertion equality is replaced by two inequalities ($\geq$ and $\leq$). 
While we demonstrate one specific transformation for each modality in our work to compute the respective Direct Effects, we note that \ModalSCM{} can be extended to any other transformations, provided that i) the mediator variables remain unchanged, and ii) the transformations are independent of the input prompt. 
In addition to DE experiments in Section~\ref{sec:experiments}, we show DE computation with one additional transformation in Appendix~\ref{sec: additional transformation}.
We use simple prefix/suffix transformations to ensure independence between variables like $S$ and $DS$ or $C$ and $DC$, to avoid correlation introduced by common transformations such as back-translation for $NL$.

\section{Experiments}
\label{sec:experiments}
\subsection{Settings}
\paragraph{Datasets.}
To select evaluation datasets, we considered three key requirements: i) the dataset should contain  code and natural language components (and preferably I/O pairs), iii) it should provide test cases to quantify code correctness, and iii) input modalities should be separable to isolate modal causal effects.

Based on these criteria, we study the causal effects on codeLLMs across three code generation benchmarks HumanEval+ \cite{evalplus}, mMBPP+ \cite{evalplus}, and CoderEval\cite{codereval}. To accommodate the lack of an explicit $Code_{AL}$ modality in the original MBPP+ dataset, we create mMBPP+ (multi-modal MBPP+) by adding a code function header to the original prompt. We evaluate HumanEval and mMBPP using evalplus \cite{evalplus}, which extends original datasets by incorporating additional challenging test cases for more rigorous testing. CoderEval offers a range of coding problems, from self-contained functions, to more complex functions that require an entire project environment to run. 
We focus on the self-contained (SC) subset, named CoderEval-SCP for Python and CoderEval-SCJ for Java. Detailed statistics are shown in Table~\ref{tab:dataset_stats}. 

\begin{table}[t]
  \centering
    \begin{adjustbox}{width=0.99\columnwidth}
  \begin{tabular}{lccccc}
    \hline
    Dataset      & Size & NL        & $Code_{AL}$   & $Code_{NL}$   & I/O Pairs    \\
    \hline
    HumanEval+   & 164  & \checkmark & \checkmark & \checkmark & \checkmark \\
    MBPP+        & 399  & \checkmark & $\times$    & \checkmark & \checkmark \\
    mMBPP+       & 399  & \checkmark & \checkmark & \checkmark & \checkmark \\
    CoderEval    & 460  & \checkmark & \checkmark & \checkmark & $\times$ \\
    CoderEval-SCP & 35   & \checkmark & \checkmark & \checkmark & $\times$ \\
    CoderEval-SCJ & 55   & \checkmark & \checkmark & \checkmark & $\times$ \\
    \hline
  \end{tabular}
\end{adjustbox}
   \caption{\label{tab:dataset_stats}Statistics and prompt modalities of HumanEval+, MBPP+, mMBPP+, and CoderEval datasets.}
\vspace{-7pt}
\end{table}

\paragraph{Models.} 
\label{sec: models}
Using \ModalSCM, we evaluate the causal effects on three codeLLMs: OpenAI GPT-4 Turbo \cite{gpt4t} (updated on January 25, 2024), 
WizardCoder-15B \cite{luo2023wizardcoder}, and Llama-3-8B \cite{llama3}. 
To further explore the implications of modal alignment with code training, we examine the modal-representation space of CodeLLaMa-13B and LLaMa-2 13B to isolate the effects of \MultiModal{} training, keeping other parts of the training process and model architecture constant. 

\paragraph{Implementation.}
\label{sec: implementation}
Following previous works on code generation \cite{chen2021evaluating}, we use the change in mean $pass@1$ accuracy ($Pr(Y=1)$) to quantify the direct and total effects after interventions on \ModalSCM{}.  
All datasets used are evaluation-only subsets, with no training involved in our experiments. For inference on all LLMs, we use a temperature of 0.01, a top\_p value of 0.95, and a batch size of 8. The open-source model experiments were conducted on a single A100 GPU with 40 GB VRAM and each run took less than 2 GPU hours. 
During experiments with self-contained CoderEval functions in Python and Java, we ensured that the transformations were equivalent across both languages.

\begin{table*}[!t]
\centering
\begin{adjustbox}{width=0.85\textwidth}
    \renewcommand{\arraystretch}{1} 

    \begin{tabular}{l|l|cc|cc|cc|cc|cc}
    \hline
    \multirow{2}{*}{Model} & \multirow{2}{*}{Modality} & \multicolumn{2}{c|}{HumanEval+} & \multicolumn{2}{|c|}{mMBPP+} & \multicolumn{2}{|c|}{CoderEval-SCP} & \multicolumn{2}{|c|}{CoderEval-SCJ} & \multicolumn{2}{|c}{Mean}\\
    
    & & TE & DE & TE & DE & TE & DE & TE & DE & TE & DE\\
    \hline
    \multirow{4}{*}{GPT-4T}& Full & \multicolumn{2}{|c|}{81.71} &\multicolumn{2}{|c|}{72.68} & \multicolumn{2}{|c|}{48.57} & \multicolumn{2}{|c|}{43.64} & \multicolumn{2}{|c}{61.65}\\
    \cmidrule{2-12}
    
    &NL & \textbf{42.08} & 1.22 & 19.05 & 4.26 & \textbf{20.00} & \textbf{2.86}* & 3.64 & 1.82 & 21.19 & 3.64 \\
    
    &$Code_{AL}$ & 1.83 & 1.22 & 1.25 & 4.01 & 8.57 & 0.00 & \textbf{43.64} & \textbf{18.18}* & 13.82 & 5.86 \\
    
    &$Code_{NL}$ & 18.91 & 1.83 & \textbf{42.86} & 2.76 & 0.00 & \textbf{2.86*} & 1.82 & 0.00 & 15.90 & 1.52 \\
    
    &I/O Pairs & 5.49 & \textbf{2.44} & 12.28 & \textbf{6.26} & N/A & N/A & N/A & N/A & 8.89 & 4.35 \\
    \hline
    
    \multirow{4}{*}{WizCoder}& Full & \multicolumn{2}{|c|}{53.05} &\multicolumn{2}{|c|}{52.63} &\multicolumn{2}{|c|}{37.14} &\multicolumn{2}{|c|}{47.27} & \multicolumn{2}{|c}{47.52}\\
    \cmidrule{2-12}
    &NL & \textbf{30.49} & 5.49 & \textbf{13.53} & 0.50 & 5.71 & \textbf{8.57*} & 10.91 & \textbf{3.64} & 15.16 & 3.70 \\
    
    &$Code_{AL}$ & 4.27 & 9.76 & 2.00 & \textbf{2.50} & 2.86 & 2.86* & \textbf{45.45} & 0.00 & 13.65 & 3.78 \\
    
    &$Code_{NL}$ & 6.10 & 2.44 & 4.01 & 0.50 & \textbf{8.57*} & \textbf{8.57*} & 3.64 & 0.00 & 5.58 & 3.34 \\
    
    &I/O Pairs & 12.20 & \textbf{12.20} & 5.26 & 0.75 & N/A & N/A & N/A & N/A & 8.73 & 6.48 \\
    
    \hline
    
    \multirow{4}{*}{LLaMa-3}& Full & \multicolumn{2}{|c|}{55.49} &\multicolumn{2}{|c|}{58.64} &\multicolumn{2}{|c|}{31.43} &\multicolumn{2}{|c|}{0} & \multicolumn{2}{|c}{36.39} \\
    \cmidrule{2-12}
    
    &NL & \textbf{33.54} & 3.66 & \textbf{16.54} & 0.00 & \textbf{11.43 }& \textbf{5.71*} & 0.00 & \textbf{3.64*} & 15.38 & 3.54 \\
    
    &$Code_{AL}$ & 0.61 & 3.66 & 1.76 & 1.51 & 0.00 & 2.86* & 0.00 & 0.00 & 0.59 & 2.01 \\
    
    &$Code_{NL}$ & 10.98 & 3.05 & 6.02 & 2.01 & 8.57 & 0.00 & 0.00 & 0.00 & 6.39 & 0.98 \\
    
    &I/O Pairs & 6.10 & \textbf{4.27} & 6.27 & \textbf{2.76} &  N/A& N/A & N/A & N/A & 6.19 & 3.52 \\
 
    \hline
    \end{tabular}
\end{adjustbox}
\caption{\label{tab:causal_effects}Total Effect (TE) and Direct Effect (DE) of modalities on code generation. Pass@1 accuracy on Full prompt for each model and dataset is reported, followed by accuracy drop, indicating TE or DE. "*" denotes an increase in accuracy with the respective intervention. Bold highlights top TE and DE for each dataset and model. Accuracy results are averaged across 3 runs.} 
\vskip -0.2in
\end{table*}

\subsection{Total Effects of Modalities}
{\bf Natural Language.}
The NL component, often a docstring in code completion tasks and containing the core logic of the generated code, shows the highest TE across all models on HumanEval+, mMBPP+, and CoderEval-SCP. As shown later in Section~\ref{sec:errors_in_codellms}, removing NL increases the semantic errors. The TE of NL is highest for HumanEval+, followed by mMBPP+ and CoderEval-SCP, likely due to the greater detail in HumanEval+ docstrings compared to the shorter ones in CoderEval-SCP. However, the $do(X_{NL}=-1)$ intervention still maintains a non-zero accuracy. Given that generating correct code output without NL semantics should not be possible, we hypothesize that the model either infers the correct NL semantics from $Code_{NL}$ or relies on its memory, suggesting memorization.

\textbf{\( \boldsymbol{Code_{NL}\ \   TE.} \)}
The latter hypothesis is confirmed by the TE computation of $Code_{NL}$, which emerges as an important prompt component in the HumanEval+ and mMBPP+ datasets. For GPT-4T on mMBPP+, the TE of $Code_{NL}$ is 42.86\%, surpassing the NL modality. Natural language chat models, such as GPT-4T and LLaMa-3, consistently show higher TE for $Code_{NL}$, with GPT-4T reaching 18.91\% on HumanEval+. This suggests that natural language models may prioritize NL semantics ($M_{NL}$) more than code-focused models.

\textbf{\( \boldsymbol{Code_{AL}\ \   TE.} \)}
In CoderEval-SCJ, $Code_{AL}$ has a high TE across all models, with GPT-4T and WizardCoder performance dropping nearly to zero. We observe limited code generation capabilities in the Java programming language exhibited by codeLLMs, particularly evident with zero performance from LLaMa-3. Further, models hallucinate code entry points when $Code_{AL}$ is absent. For instance, in all 55 examples, LLaMa-3 places the required code in a hallucinated class, as illustrated in Figure~\ref{fig:naacl_codeal_ce_inc}. On Python subsets, $Code_{AL}$, which contains minimal syntax information such as function headers and variable names, has the lowest TE across all models. However, $Code_{AL}$ in all three datasets under consideration is limited to the function header and input variable names along with function syntax (Figure~\ref{fig:prompt_example}); the TE of $Code_{AL}$ where it may contain essential generation logic is yet to be explored.

{\bf I/O Pairs.}
The TE of I/O pairs surpasses that of $Code_{NL}$ with WizardCoder and holds equal significance with LLaMa-3 and GPT-4. This underscores the syntactic information encoded within I/O pairs, potentially aiding the model in reasoning over correct code structures. Analogously to human programmers employing unit testing for iterative code writing, the TE of I/O pairs suggests a similar process within codeLLMs. Future versions of codeLLMs may leverage intermediate I/O values for handling complex code, similar to the debugging process in software engineering.

\begin{figure*}[!t]
    \centering
    \begin{subfigure}{0.24\linewidth}
        \centering
        \includegraphics[width=\linewidth]{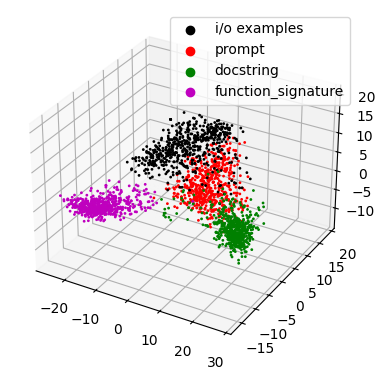}
        \caption{\label{subfig1}CodeLLaMa-13B}
    \end{subfigure}
    \begin{subfigure}{0.24\linewidth}
        \centering
        \includegraphics[width=\linewidth]{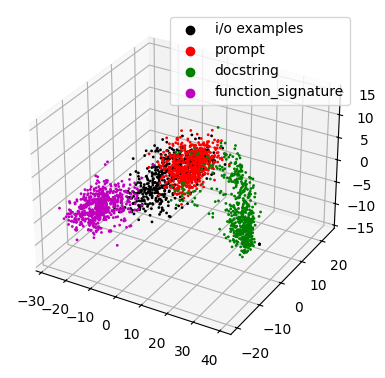}
        \caption{\label{subfig2}LLaMa-2 13B}
    \end{subfigure}
    \begin{subfigure}{0.24\linewidth}
        \centering
        \includegraphics[width=\linewidth]{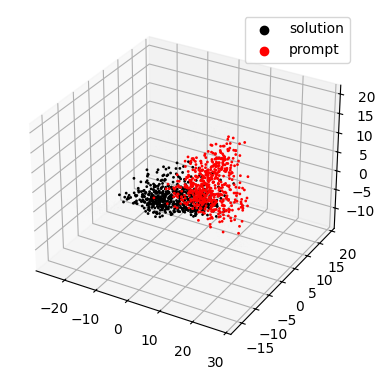}
        \caption{\label{subfig3}CodeLLaMa-13B}
    \end{subfigure}
    \begin{subfigure}{0.24\linewidth}
        \centering
        \includegraphics[width=\linewidth]{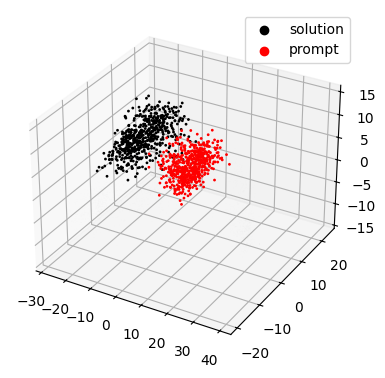}
        \caption{\label{subfig4}LLaMa-2 13B}
    \end{subfigure}
    \caption{(a) and (b) Embedding PCA projections of modalities in input prompt by CodeLLaMa and LLaMa-2. (c) and (d) Prompt embedding projections along with the ground-truth code embedding projections by CodeLLaMa and LLaMa-2. $Code_{AL}$ and $Code_{NL}$ is combined into function\_signature.}
    \label{embedding-isotropy-figure}
\end{figure*}

{\bf Memorization of Code Benchmarks.}
Given that codeLLMs are trained on open-source datasets, we explore the potential for benchmark memorization. The non-zero $pass@1$ accuracy, even without NL instructions, indicates memorization. Furthermore, even after standardizing function header names, GPT-4T still generated original function names in 11.5\% of HumanEval+ and 7.2\% of mMBPP+ cases (Figure \ref{fig:naacl_codeal_ce_inc}). LLaMa-3 showed similar behavior, with 10.3\% of HumanEval+ examples despite the standardization of function names. The notably high memorization figures for GPT-4T also raise concerns regarding its performance on the EvalPlus leaderboard\cite{evalplus}. Similar to prior studies, such as \cite{Lai2022DS1000}, that examine memorization, our causal analysis also suggests substantial dataset memorization. However, a detailed investigation is left for future work.

\subsection{Direct Effects of Modalities}
\label{sec:direct effects}
\begin{figure*}[t]
    \centering
    \includegraphics[width=0.99\textwidth]{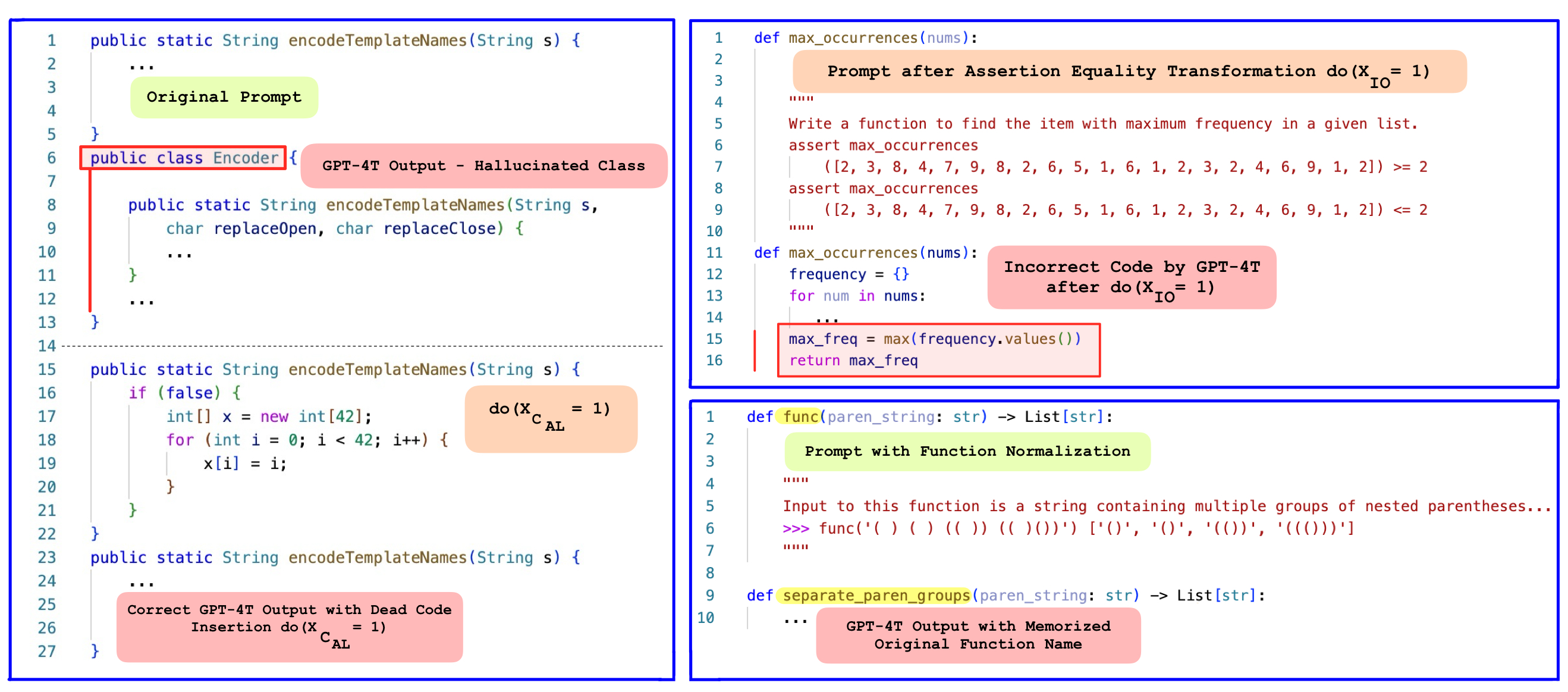}
    \caption{\label{fig:naacl_codeal_ce_inc}Left figure shows a CoderEval-SCJ prompt where dead code insertion corrects the original prompt's error of creating a hallucinated Java class (\textcolor{red}{red} box). The top right figure illustrates an mMBPP+ prompt where  I/O pair transformations lead to a semantic error in lines 15-16. The bottom right figure shows GPT-4T’s memorization of a HumanEval+ prompt.}
\vspace{-15pt} 
\end{figure*}

We define direct effects (DE) by noting the drop in $pass@1$ accuracy of the model under the semantics-preserving transformations of modalities where the latent mediators remain unchanged (Section~\ref{sec:causal effects defn}). These effects also represent the spurious correlations, as any non-spurious learning process must be mediated through $M_{NL}$ and $M_{Code}$. From Table~\ref{tab:causal_effects}, I/O pairs exhibit the strongest direct effect (DE) on HumanEval+ and mMBPP+ across models, except for mMBPP+ on WizardCoder. As seen in Figure \ref{fig:naacl_codeal_ce_inc}, replacing a single assert equality in each I/O example with two inequalities makes it harder for the model to reason correctly over the code logic.

The DE of I/O pairs is then followed by the DE of $Code_{AL}$, where WizardCoder shows a very high DE of 9.76\% on HumanEval+. For CoderEval-SCJ, GPT-4T's accuracy increased by 18.18\% under the $do(X_{AL}=1)$ intervention. As shown in Figure \ref{fig:naacl_codeal_ce_inc}, a Java code snippet in the form of dead code reduces the class name hallucinated by the model. With this finding, we speculate that dead code might help control hallucinations, but we leave the detailed analysis to future work. In general, we observe that the DEs of $NL$ and $Code_{NL}$ are comparatively lower, implying models are more robust to natural language than code semantics, likely due to instruction tuning stages.

\begin{table*}[t]
\centering
\begin{adjustbox}{width=0.7\textwidth}
    \begin{tabular}{l|l|cccc|cccc}
    \hline
     & & \multicolumn{4}{c|}{HumanEval+} & \multicolumn{4}{c}{mMBPP+} \\
    Model & Modality & Syn & Sem & Runt & Other & Syn & Sem & Runt & Other \\
    \hline

    \multirow{4}{*}{GPT-4T}& 
    NL & -8.22 & 26.13 & -18.67 & 0.76
        & -5.58 & 14.32 & -9.61 & 0.87 
          \\
    
    &$Code_{AL}$ & -0.75 & 1.49 & -1.80 & 1.10 
    & -1.42 & 2.31 & -0.89 & 0.00 
     \\
    
    &$Code_{NL}$ & -3.12 & -7.04 & -8.47 & 18.63 
    & -11.27 & -30.64 & 32.21 & 9.70  \\

    &I/O Pairs & -1.62 & 0.61 & 1.01 & 0.00
        & -3.85 & -4.75 & 7.65 & 0.94\\

    \hline

    \multirow{4}{*}{WizCoder}& NL & 
        -3.37 & 15.01 & -11.65 & 0.00
        & -1.29 &7.72& -6.01 &-0.42 \\
    
    & $Code_{AL}$ &
        -0.30 & -0.65 & -0.95 & -.00
        & -0.92& 6.93& -5.58& -0.42\\

    & $Code_{NL}$ & 
    -0.59 & -4.58 & -0.21 & 4.96 &
    -1.04 & -2.08 & 2.74&  0.39\\

    & I/O Pairs
    & -1.01 & 2.67 & -1.65 & 0.0 &
    -1.05 & 4.06 & -3.38 & 0.37\\    

    \hline

    \multirow{4}{*}{LLaMa-3}
    & NL 
    & -4.11 & 20.96 & -16.60 & -0.25 
    & -5.14 & 14.13 & -7.98 & -1.01
    \\
    
    &$Code_{AL}$
    & 0.42 & 0.90 & -1.30 & -0.03 
    & -1.32 & 4.38 & -2.13 & 0.93
    \\

    &$Code_{NL}$ 
    & -1.57 & -8.67 & -1.21 & 11.44 
    & -0.90 & -10.04 & 7.68 & 3.26
    \\

    &I/O Pairs 
        & 4.43 & 1.75 & -6.13 & -0.05
        & -0.41 & -9.85 & 10.80 & -0.54
        \\
    
    \hline
    \end{tabular}
\end{adjustbox}
\caption{\label{tab:combined_errors_gpt} Percentage of errors out of total passed cases for GPT-4T, WizardCoder-15B, and Llama-3-8B on HumanEval+ and mMBPP+. Negative percentages indicate a decrease in error count, while positive values indicate an increase in error count upon intervention. Syn represents Syntax errors, Sem represents Semantic Errors, and RunT represents runtime errors.}  
\end{table*}

\subsection{Effect of Multi-Modal Pretraining}
\label{sec: multimodal pretraining}
Our experiments on causal effects reveal asymmetric impacts of different modalities, leading us to examine their distribution in the embedding space of codeLLMs with code pretraining. We use PCA, following previous works \cite{cai2020isotropy, rajaee2021isotropy}, to visualize high-dimensional representations into three dimensions. For this analysis, we combine data samples from the HumanEval and mMBPP datasets, excluding CoderEval due to its lack of the I/O modality (Table~\ref{tab:dataset_stats}). 

We use LLaMa-2\cite{touvron2023llama2} to explore the effect of multi-modal pretraining of LLMs and how it affects the embedding representation of different modalities. The pretraining stages of code-aware LLMs add multi-modal alignment in codeLLMs, which is confirmed by the performance difference of 26.9\% on HumanEval and 39.2\% on MBPP between the LLaMa-2 (13B) and the codeLLM CodeLLaMa (13B), as reported by prior works \cite{li2023starcoder, liu2023code}. Using CodeLLaMa-13B and LLaMa-2 13B we can isolate the effects of multi-modal training keeping other factors such as model architecture and positional encodings constant.

In Figure~\ref{subfig2}, complete prompts and modal components (red and green) form distinct clusters in the case of LLaMa-2, whereas, in Figure~\ref{subfig1}, the prompt and the docstring are better associated by CodeLLaMa as they form closer clusters. I/O examples in the prompt have high token overlap with the function header (Figure~\ref{fig:prompt_example}). While LLaMa-2 keeps them together (black and magenta clusters) probably due to the high token overlap, CodeLLaMa can separate them in the embedding space. Figures~\ref{subfig3} and \ref{subfig4} show how the model associates input prompts and ground-truth code solutions. LLaMa-2 forms nearly two disjoint clusters (Figure~\ref{subfig3}), even when prompts and ground-truth code are strongly correlated, while CodeLLaMa can associate prompt with code solution in the embedding space. We further discuss anisotropy in CodeLLaMa's embedding spaces in Appendix~\ref{app:\MultiModal{}-pretraining}.

\subsection{Errors in codeLLMs}
\label{sec:errors_in_codellms}
In this section, we analyze the types of errors encountered by codeLLMs on the HumanEval+ and mMBPP+ datasets, as detailed in Table~\ref{tab:combined_errors_gpt}. Errors are categorized into syntax errors, semantic errors, runtime errors, and other errors. 
The values in the table represent the percentage change in error counts upon removal of the modality ($do(X=-1)$) during TE computation, relative to the full prompt. For instance, upon removal of $NL$ from HumanEval for GPT-4T leads to a drop in accuracy of 42.08\% (Table~\ref{tab:causal_effects}), and the 26.13\% of this drop is due to increase in semantic errors (Table~\ref{tab:combined_errors_gpt}). So the percentage changes sum up to zero for each modality, model, and dataset combination. As evident for all three models, removing the NL modality leads to a significant increase in semantic errors, confirming that natural language instructions are crucial for conveying problem semantics. Semantic errors are seen in form of failed test cases or assertion statements.
Other errors such as resource, dependency, environment and timeout errors are mostly seen with the removal of $Code_{NL}$ modality, reflecting its importance in guiding the correct code syntax.
WizardCoder-15B shows relatively small changes in syntax and other errors across modalities. For instance, on HumanEval+, syntax errors change by -0.30\% to -3.37\% across different modalities, indicating strong syntactic generation capabilities, likely due to its extensive code-specific training.

\section{Related Work}
\paragraph{Automatic Code Generation.}
Code generation with multi-modal prompts has been explored by some earlier works such as \cite{desai2015program, gulwani2017program}. Recent works have either adopted the transformer architecture \cite{feng2020codebert, wang2021codet5} or leveraged the GPT \cite{brown2020language} skeleton with massive pretraining for code pretraining \cite{rozière2023code, li2023starcoder, luo2023wizardcoder, nijkamp2023codegen, zhu2024deepseek}.

\paragraph{Prompt-Tuning.} 
Various approaches to prompt-tuning \cite{white2023prompt} have been explored for various domains and modalities \cite{mullick2024intent, wu2023self}, such as Chain-of-Thought reasoning \cite{wei2023chainofthought}, Tree of Thoughts \cite{yao2023treethought}, discrete prompt optimization \cite{wen2023hard, shin2020autoprompt}, and few-shot learning \cite{brown2020language}. In the context of code generation, prompt engineering has been leveraged for human-in-loop debugging \cite{denny2022conversing}, correctness evaluation of generated code \cite{liu2023code}, multistep planning, and generation \cite{zheng2023outline}. Our work explores the effects of modalities in prompts on code generation, which can be further used for targeted prompt-tuning processes for better performance.

\paragraph{Causal Inference in Code/NLP.} 
Recent research has applied causal inference to the NLP domain \cite{vig2020investigating, finlayson2021causal, stolfo2022causal} to better understand model behavior, which is now formalized as causal NLP \cite{jin-etal-2022-causalnlp, feder2022causal}. 
In the context of code, prior approaches have applied causal framework for various classification tasks such as vulnerability detection \cite{rahman2024towards, 9793858} and code performance prediction \cite{cito2021counterfactualexplanationsmodelscode}.
To the best of our knowledge, we are the first to apply causal inference to study modal effects on code generation task.

\section{Conclusion}
\label{sec:conclusion}
We propose \ModalSCM{}, a Structural Causal Model for analyzing \MultiModal{} code generation using LLMs. Our analysis revealed that input-output examples and natural language code components significantly influence model generation. Additionally, our interventions show that semantics-preserving changes can impact accuracy and can also lead to fewer hallucinations in some cases.

\section{Limitations} We can calculate causal effects in \ModalSCM{} with the assumption of no confounders. We believe that in the future, our causal formulation of code generation could be extended to account for confounders using the backdoor criterion.

\section{Ethical considerations} 
We release the algorithmic details and work with public code datasets, which neither reveal any personal sensitive information nor contain any toxic statements. While there are potential societal consequences, they are not deemed important to highlight here.

\section{Acknowledgments} 
We thank the anonymous reviewers for their constructive and valuable feedback. This work is supported partially by an NSF CAREER award; and an award from the Google Cyber NYC Institutional program. Any opinions, findings, conclusions, or recommendations expressed herein are those of the authors and do not necessarily reflect those of NSF, or Google.

\bibliography{acl_latex}

\appendix

\section{Causal Effects}
\label{app:causal_effects_extended}
In this section, we present the causal effects of three modalities: Natural Language (NL), I/O Pairs, and Code with NL component (\(Code_{NL}\)). For each modality, we provide the corresponding structural equation, followed by the total effect (TE) and direct effect (DE).

\subsection{Natural Language (NL)}
\label{sec:nl_effects}
The $NL$ variable is defined by the following structural equation:
\begin{multline}
\label{eq:nl intervention}
    NL \gets \mathbbm{1}_{\{ X_{NL} = 1 \}}(S + DS)\\ + \mathbbm{1}_{\{ X_{NL} = 0 \}}(S) + \mathbbm{1}_{\{ X_{NL} = -1 \}}(NULL)
\end{multline}
where $S$ is the actual natural language prompt component $\mathcal{P}_{S}\in \mathcal{D}$, $DS$ is a Dead String that does not alter the semantics of the natural language, and $X_{NL}$ is used to control whether to allow the original $\mathcal{P}_{S}$, concatenate a dead string, or remove the natural language modality. Similar to dead code insertion in $Code_{AL}$, dead string insertion is a semantics-preserving transformation such that $M_{NL}(S) = M_{NL}(S + DS)$. $NL$ subscript is dropped for brevity.

\paragraph{Total Effect of NL.}
\begin{align*}
    TE &= TE(do(X_{NL}=0), do(X_{NL}=-1))\\
    &= Acc(\mathcal{D}) - Acc(\mathcal{D}; \mathcal{P}_{NL} = NULL)
\end{align*}

\paragraph{Direct Effect of NL.}
\begin{align*}
    DE &= \mathbb{E}\left[Y_{X=1, NL(X=1), M_{NL}\left( NL(X=0) \right)}\right]\\
    &\quad- \mathbb{E}\left[Y_{NL(X=0)}\right]\\
    &= Acc(\mathcal{D}) - Acc(\mathcal{D}; \mathcal{P}_{NL} = S + DS)
\end{align*}
Here, $DS$ represents the Dead String. We use the prefix `DOCSTRING: ' concatenated to each natural language instruction to preserve semantics. Other transformations such as back-translation are possible but introduce correlations between variables, so we prefer simpler prefix or suffix transformations that keep $S$ and $DS$ independent.

\subsection{I/O Pairs}
\label{sec:io_effects}

The I/O modality is defined by the following structural equation:
\begin{multline}
\label{eq:io intervention}
    I/O \gets \mathbbm{1}_{\{ X_{IO} = 0 \}}(I^r = I^r)\\ + 
    \mathbbm{1}_{\{ X_{IO} = 1 \}}((I^l \leq I^r) + (I^r \geq I^r))\\ + 
    \mathbbm{1}_{\{ X_{IO} = -1 \}}(NULL)
\end{multline}
where $I^l$ and $I^r$ represent the left-hand side (LHS) and right-hand side (RHS) of the assertion equality statement in the original prompt, respectively. For semantics-preserving transformations, we replace each assertion equality with two inequalities, $\leq$ and $\geq$. $I/O$ is omitted for brevity.

\paragraph{Total Effect of I/O.}
\begin{align*}
    TE&= TE(do(X_{IO}=0), do(X_{IO}=-1))\\
    &= Acc(\mathcal{D}) - Acc(\mathcal{D}; \mathcal{P}_{IO} = NULL)
\end{align*}

\paragraph{Direct Effect of I/O.}
\begin{align*}
    DE &= \mathbb{E}\left[Y_{\left(X=1, M_{Code}(X=1),M_{Code}(X=0) \right)} \right]\\
    &\quad\quad - \mathbb{E}\left[Y_{M_{Code}(X=0)}\right]\\
    &= Acc(\mathcal{D}) - \\
    &\quad\quad Acc(\mathcal{D}; \mathcal{P}_{IO} = (I^l \leq I^r) + (I^r \geq I^r))
\end{align*}

\subsection{Code with NL Component (\(Code_{NL}\))}
\label{sec:codenl_effects}
The $Code_{NL}$ modality is defined by the following structural equation:
\begin{multline}
\label{eq:dual intervention}
    Code_{NL} \gets \mathbbm{1}_{\{ X_{NL} = 1 \}}(C_{NL} + DN)\\ 
    + \mathbbm{1}_{\{ X_{NL} = 0 \}}(C_{NL}) + \mathbbm{1}_{\{ X_{NL} = -1 \}}(NULL)
\end{multline}
where $C_{NL}$ is the code prompt component $\mathcal{P}_{Code_{NL}}\in \mathcal{D}$, and $DN$ is a Dead Name added to the function header. This transformation preserves semantics for both the natural language and code domains. For instance, $M_{NL}(C_{NL}) = M_{NL}(C_{NL} + DN)$.

\paragraph{Total Effect of \(Code_{NL}\).}
\begin{align*}
    TE&= TE(do(X_{CN}=0), do(X_{CN}=-1))\\
    &= Acc(\mathcal{D}) - Acc(\mathcal{D}; \mathcal{P}_{Code_{NL}} = NULL)
\end{align*}

\paragraph{Direct Effect of \(Code_{NL}\).}
\begin{align*}    
    DE &= \mathbb{E}\left[Y_{\left(X=1, C_{NL}(X=1), M_{NL}\left( C_{NL}(X=0) \right), \right.}\right.\\
    &\quad \left._{\left. M_{C}\left(C_{NL}(X=0) \right)\right)} \right] - \mathbb{E}\left[Y_{(C_{NL}(X=0)}\right]\\
    &= Acc(\mathcal{D}) - Acc(\mathcal{D}; \mathcal{P}_{C_{NL}} = C_{NL} + DN)
\end{align*}
Here, $DN$ represents Dead Name, and we use the prefix `func\_' in Python and `Method' in Java to maintain semantic preservation. Other transformations, like capitalization, are possible but avoided to keep $C_{NL}$ and $DN$ independent.

\section{Multi-Modal Prompt} \label{appendix
}
The \MultiModal{} prompt $\mathcal{P}$ can be expressed as an equation comprising one or more prompt components $P^{j}$ of modality $M_i$, where different prompt components are concatenated using one of the defined separators: 
\begin{align}
\label{eq:mm_prompt}
&\mathcal{P} = P^{1}_{M_1} \left[ \text{sep} \ \ P^{j}_{M_i} \right]\\ &\text{sep} = \ '\ \ ' \ \ | \ \ \symbol{92}n \ \ | \ \ \symbol{92}t \ \ | \ : \ \ | \ \ , \ \ | \ ; 
\end{align}
In this equation, different prompt components are concatenated using one of the defined separators.


\section{Implementation Details}
\label{app:implementation_details}
We exclude APPS \cite{hendrycksapps2021} and CodeContest \cite{li2022competition}, as they lack \MultiModal{} prompts, making them unnecessary for \MultiModal{} causal analysis. Similarly, while the CONCODE segment of the CodexCGLUE \cite{lu2021codexglue} benchmark includes \MultiModal{} prompts, it measures code quality via natural language similarity metrics like BLEU, which is unsuitable for code generation tasks. Lastly, DS-1000 \cite{Lai2022DS1000} was excluded due to the need for manual screening of all examples to separate modal components for \ModalSCM{}.



\section{DE Additional Transformation}
\label{sec: additional transformation}
We demonstrate one specific transformation for each modality in the paper and compute the respective causal effects. CodeSCM can be directly extended to other transformations as well for DE computation. For example, in Table~\ref{tab: additional de table}, along with original transformations from Table~\ref{tab:causal_effects} (DE-1), we illustrate DE computation with an additional set of transformations (DE-2) for the mMBPP+ dataset using WizardCoder codeLLM. The following transformations are used for DE-2 - (dead string prefix, unused variable, dead name prefix, and negating the not assert statement):
\begin{itemize}
    \item $DS$ = Code Logic:\symbol{92}n (in Equation~\ref{eq:nl intervention})
    \item $C_{DC}$ = \symbol{92}tvar = 42 (in Equation~\ref{eq:code intervention})
    \item $DN$ = header\_ (in Equation~\ref{eq:dual intervention})
    \item assert $I^l == I^r$ is changed to assert not $I^l != I^r)$ (in Equation~\ref{eq:io intervention})
\end{itemize}

\begin{table}[h]
  \centering
  \begin{tabular}{l|ll}
    \hline
    Modality & DE-1 & DE-2\\
    \hline
    Full & \multicolumn{2}{|c}{52.63} \\
    \hline
    NL & 0.50 & 1.23\\
    $Code_{AL}$ & 2.50 & 3.03\\
    $Code_{NL}$ & 0.50 & 1.73\\
    I/O Pairs & 0.75 & 3.23\\
    \hline
  \end{tabular}
  \caption{\label{tab: additional de table}Direct effects of WizardCoder on mMBPP+ dataset under an additional transformation. DE-1 values are the same as Table~\ref{tab:causal_effects}}
\end{table}

\section{Multi-Modal Pretraining}
\begin{table}[!h]
\centering
\begin{adjustbox}{width=0.99\columnwidth}
    \begin{tabular}{llcc}
    \hline
    Prompt Component & CodeLLaMa $\downarrow$ & LLaMa-2 \\
    \hline
examples & 0.85 & 0.77 \\
docstring & 0.86 & 0.83 \\
prompt & 0.87 & 0.80 \\
solution & 0.90 & 0.83 \\
function & 0.91 & 0.85 \\
\hline
all & 0.77 & 0.66 \\ 
    \hline
    \end{tabular}
\end{adjustbox}
\caption{\label{app:\MultiModal{}-pretraining}Intra-modal cosine similarity between mean hidden representation of CodeLLaMa-13B and LLaMa-2. Similarities are reported by combining HumanEval and mMBPP.}
\end{table}

\begin{table}[!h]
\centering
\begin{adjustbox}{width=0.99\columnwidth}
    \begin{tabular}{llcc}
    \toprule
    Modality-1 & Modality-2 & CodeLLaMa $\downarrow$ & LLaMa-2 \\
    \midrule
function & docstring & 0.59 & 0.43 \\
docstring & examples & 0.63  & 0.45\\
solution & docstring & 0.65  & 0.47 \\
function & prompt & 0.74  & 0.66 \\
docstring & prompt & 0.76  & 0.64\\
function & examples & 0.77  & 0.66  \\
examples & prompt & 0.82  & 0.72  \\
solution & function & 0.82  & \textbf{0.76}  \\
solution & examples & 0.83  & 0.72  \\
solution & prompt & 0.84  & 0.72  \\
    \bottomrule
    \end{tabular}
\end{adjustbox}
\caption{\label{tab:inter_modal}Inter-modal cosine similarity between averaged hidden representation of CodeLLaMa-13B and LLaMa-2. Similarities are reported by combining HumanEval and mMBPP.}
\end{table}

Inspired from previous works \cite{cai2020isotropy, rajaee2021isotropy}, we measure the cosine similarities between the averaged last layer's hidden state representations of CodeLLaMa and LLaMa-2. For each modality, intra-modality $S_{intra}$ cosine similarity is defined as $\mathbb{E}_{i,j\in P} \left[cos\left(M(i),M(j)\right)\right]$, where $i$ and $j$ are distinct prompts of same modality. Inter-modal cosine similarity $S_{inter}$ is defined for a pair for modalities, as $\mathbb{E}_{i\in P1,j\in P2} \left[cos\left(M(i),M(j)\right)\right]$, where $i$ and $j$ belong to different prompt modal components. 

Similar to Section~\ref{sec: multimodal pretraining}, we combine data samples from the HumanEval and mMBPP datasets, excluding CoderEval due to its lack of I/O modality (Table~\ref{tab:dataset_stats}). In Table \ref{tab:inter_modal}, the ground truth problem solution and input prompt are kept closest by CodeLLaMa despite being of different modalities and low token overlap, which explains CodeLLaMa's superior performance on code generation benchmarks. LLaMa-2 on the other hand, keeps ground truth problem solution and function header name, probably due to significant token overlap between the two.

Measuring $S_{inter}$ for each modality, in Table \ref{tab:inter_modal} we observe the closer clusters of modalities in CodeLLaMa's vector space with consistently higher similarities. Given a round of code pretraining, CodeLLaMa assigns the highest $S_{inter}$ to the function header and solution, both of which are code components, while LLaMa-2 assigns similar $S_{inter}$ to the docstring ($P_{NL}$) and code solution ($P_{code}$). Finally, in the last row of Table \ref{tab:inter_modal}, we show the average cosine similarity of the entire space i.e., vector representations from all components. We note a higher similarity in the code model. 
Given elevated values of similarity for the code model, we suspect an anisotropic embedding space compared to the natural language model. Anisotropy would increase as the model learns to specialize in one task/domain (code generation in this case) and loses generalization capabilities. The concrete conclusive claim however requires further analysis which we leave to the future works.

\label{sec:types_of_errors}

\section{Error Analysis}
\label{app:types_of_errors}

In this section, we provide definitions of the different types of errors encountered in code generation tasks.

\paragraph{Syntax Errors.} These errors occur when the code does not conform to the syntactical rules of the programming language. They are typically detected during the parsing stage. An example of a syntax error might be a missing colon, unmatched parentheses, or incorrect indentation.

\paragraph{Semantic Errors.} Semantic errors arise when the code is syntactically correct but fails to produce the intended output due to logical mistakes. This can include errors in the logic of the code, incorrect use of variables, or wrong implementation of algorithms. We broadly encounter two types of semantic errors: (i) \textbf{test case errors}, when the test cases in the respective dataset fail; (ii) \textbf{assertion errors}, when an input-output example assertion in the prompt fails.

\paragraph{Runtime Errors.} These errors occur during the execution of the code. They result from operations like division by zero, accessing out-of-bound indices, or other exceptional conditions that the code does not handle.

\paragraph{Other Errors.} This category includes various errors that do not fit into the above classifications. It covers resource errors (e.g., memory errors when the program tries to allocate more memory than what is available), dependency errors (e.g., missing modules or packages), environment errors (e.g., issues with file access or permissions), and timeout errors (when the execution of the code takes longer than the allowed time limit).

\end{document}